\documentclass{article}
\usepackage{spconf,amsmath,amssymb,graphicx}
\usepackage[T1]{fontenc}
\usepackage[utf8]{inputenc}
\usepackage{booktabs,multirow,makecell}
\usepackage{stfloats}
\usepackage{url}
\usepackage[table]{xcolor}

\usepackage{enumitem}
\setlist{nosep, leftmargin=14pt}

\definecolor{cxteal}{HTML}{2A9D8F}
\definecolor{cxorange}{HTML}{E76F51}
\definecolor{cxdark}{HTML}{264653}

\newif\ifdraft \drafttrue
\ifdraft
  \newcommand{\todo}[1]{{\color{red}\textbf{[TODO:} #1\textbf{]}}}
  
\else
  \newcommand{\todo}[1]{}
  
\fi

\newcommand{\nkp}{317}        
\newcommand{\ngroups}{27}     
\newcommand{\ncand}{803}      
\newcommand{\nfinal}{132}     
\newcommand{\nslots}{311}     
\newcommand{\nkmax}{13}        

\makeatletter
\long\def\@makecaption#1#2{\vskip\abovecaptionskip
  \setbox\@tempboxa\hbox{#1. #2}%
  \ifdim\wd\@tempboxa>\hsize #1. #2\par\else\hbox to\hsize{\hfil\box\@tempboxa\hfil}\fi
  \vskip\belowcaptionskip}
\makeatother
\makeatletter
\long\def\@makecaption#1#2{%
  \vskip 3pt
  \setbox\@tempboxa\hbox{#1. #2}%
  \ifdim \wd\@tempboxa >\hsize #1. #2\par \else \hbox to\hsize{\hfil\box\@tempboxa\hfil}\fi}
\makeatother

\title{Semantic landmark detection in chest radiographs}

\name{Constantin Seibold \qquad Nils Havemann \qquad David Vinu \qquad Matthias A. Fink}
\address{Clinic for Diagnostic and Interventional Radiology,\\
Heidelberg University Hospital, Heidelberg, Germany\\
\texttt{\{firstname.lastname\}@med.uni-heidelberg.de}}

\begin{document}
\ninept
\maketitle

\begin{abstract}
Quantitative chest radiograph (CXR) readouts such as the cardiothoracic ratio,
costophrenic angles or device-tip positions are defined by anatomical landmarks,
yet there currently is no agreed set of named CXR landmarks and no annotated reference to evaluate
them. We compile a list of \nkp{} candidate landmarks from the clinical literature, each defined by a
geometric rule on anatomy masks. Two reviewers with a medical background curate it independently, followed by a board-certified radiologist deciding the final taxonomy of \nfinal{} landmarks.
Projecting the rules onto the masks of CT-derived PAXRay++ yields silver training labels for automated keypoint prediction. We release a double-read gold set on real CXRs with per-landmark inter-reader tolerances. Since every landmark is typically defined by the interplay of anatomical structures, we investigate whether segmentation and landmark detection should be learned together. Under a shared RAD-DINO encoder, adding mask supervision to a heatmap detector improves landmark OKS-AP by 0.7 {\scriptsize[0.5, 0.9]} at a segmentation cost of 0.7 AP, whereas instance-first joint models (Mask R-CNN, query-based) fall below the landmark-only detector. On real CXRs, direct prediction exceeds rules on predicted masks by 6.8 {\scriptsize[2.5, 10.2]} OKS-AP, and diagnostic criteria measured on the landmarks exceed the same criteria on CXAS masks by 0.08--0.45 mean AUROC.
\end{abstract}


\section{Introduction}
\label{sec:intro}

Much of what a radiologist reads from a chest radiograph (CXR) is defined by named
anatomical landmarks. The cardiothoracic ratio (CTR) relates the outermost cardiac
borders to the inner thoracic width, effusion
blunts the costophrenic angle~\cite{blackmore1996}, and tube and catheter
tips are judged against the carina and the cavoatrial
junction~\cite{goodman1976,baskin2008}. Findings themselves are localised the same way:
\emph{``at the right hilum''}, \emph{``above the left hemidiaphragm''}. Today, each readout is
measured by hand or by its own dedicated model. Making these points
explicit would let one model serve many readouts and tie each measurement and finding to a
named location a reader can check.

While landmark annotations for anatomy in CXRs exist, but they currently carry no semantic meaning. Established sets
sample organ contours densely to model shape~\cite{chexmask}, so their points are indexed along
a contour rather than being named landmarks relating to literature, whereas most named points appear only in single-purpose tools such as
automated CTR. Fine-grained anatomy segmentation of
CXRs~\cite{paxray,cxas} delineates structures but not the landmarks clinicians measure between
them. What is missing is a clinically agreed set of named landmarks and expert annotations on
real CXRs.

\begin{figure}[t]
  \centering
  \includegraphics[width=0.9\linewidth]{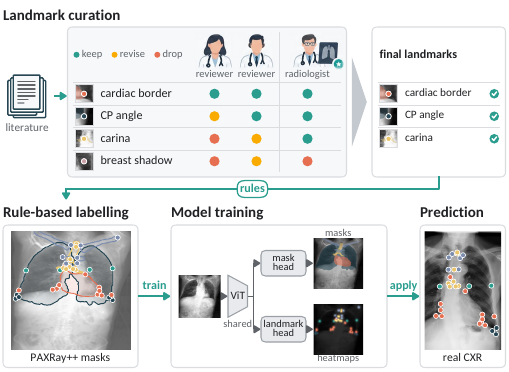}
  \caption{\textbf{Overview.} Clinicians select landmarks from literature-based candidates.
  Geometric rules on PAXRay++ masks create silver labels, on which models are trained to predict
  masks and named landmarks on real CXRs.}
  \label{fig:landmarkset}
\end{figure}

\begin{figure*}[t]
  \centering
  \includegraphics[width=\textwidth]{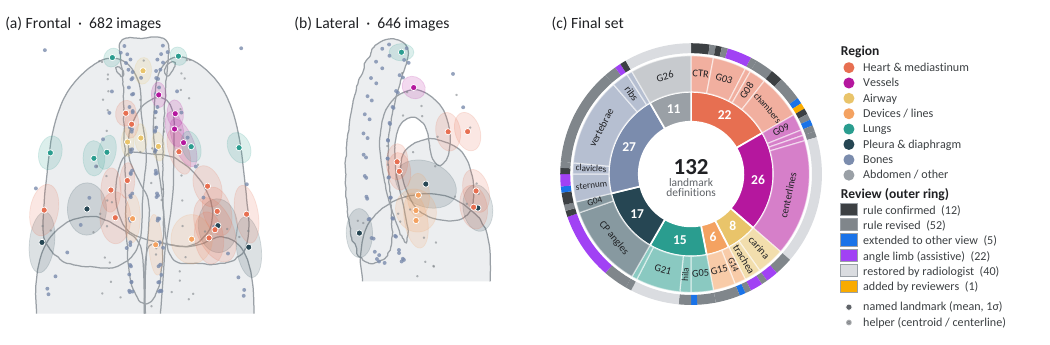}
  \caption{Final landmark set (PAXRay++ val): (a, b) mean positions, 1$\sigma$ spread; (c) definitions by region, group (G\#) and review outcome.}
  \label{fig:atlas}
\end{figure*}

To close this gap, we define each landmark as a geometric rule on anatomy masks, which gives clinicians a precise
definition to review and yields training labels wherever masks exist. Subsequently, we test whether landmarks should be predicted directly or derived by the rules from
\emph{predicted} masks, which may be not as clean on radiographs in the wild. Second, as a landmark and its
mask often describe the same underlying structure, whether both should be learned jointly, by dense or
instance-first models. 

\noindent We see our contributions are as the following:
\begin{itemize}
  \item We introduce the a comprehensive, clinically curated set of semantically
        named CXR landmarks. Its \nfinal{} landmarks were selected from \nkp{}
        literature-grounded candidates in \ngroups{} clinical groups by two independent
        reviewers with a medical background and a board-certified radiologist.
  \item We provide a gold standard of double-read landmarks on 50 MIMIC-CXR images,
        together with landmark-specific inter-reader tolerances.
  \item We benchmark rules on predicted masks, landmark-only and joint models on synthetic
        and real CXRs. We see that direct prediction is more accurate on real CXRs, and 14 diagnostic
        criteria measured on the landmarks match or exceed the same criteria on CXAS masks.
  \item We find that mask supervision slightly improves a dense landmark detector, whereas
        instance-first joint models do not reach the single-task baseline.
\end{itemize}
We release the landmark definitions, gold and silver annotations, code and model
weights at \url{https://github.com/ConstantinSeibold/cxr-landmarks}.

\section{Method}
\label{sec:method}

A CXR contains anatomical instances $\{(y_j, M_j)\}_{j=1}^{G}$ with class $y_j$ and
binary mask $M_j$. Each class $y$ has an ordered list of $K_y\le K_{\max}$ landmarks.
We denote the $k$-th landmark of instance $j$ by $\mathbf{p}_{jk}\in\mathbb{R}^2$ and
its visibility by $v_{jk}\in\{0,1\}$, and call
$\mathcal{V}=\{(y,k): k\le K_y\}$ the landmark vocabulary.


\subsection{Landmark vocabulary}
\label{ssec:taxonomy}

\noindent\textbf{Candidate list.} We compile candidate landmarks from the clinical literature,
each a geometric rule $\mathbf{p}=r(M_y,M_{y'})$ on the mask of one structure or two
adjacent ones (Fig.~\ref{fig:landmarkset}): extreme points
($\arg\max_{\mathbf{x}\in M}\mathbf{u}^{\!\top}\mathbf{x}$ for a direction $\mathbf{u}$),
contact points between structures, ridge and centerline points along boundaries and
medial axes, and centroids. On the 177 PAXRay++ structures~\cite{cxas} this gives
\nkp{} definitions and \ncand{} class--landmark pairs in \ngroups{} clinical groups,
linked to 53 sources (e.g.\ \cite{hansell2008}).

\noindent\textbf{Expert review.} Two reviewers with a medical background independently rated
every candidate for clinical relevance (five-point scale), placement (keep, revise,
wrong, merge, drop) and visibility on real CXRs. A board-certified radiologist then
decided the final list. The reviewers kept 64 definitions and proposed one (the
Hoffman--Rigler point). The radiologist restored 40 dropped geometric descriptors (lung
zones, vessel centerlines, organ centroids). We merged duplicates, removed bilateral
union classes, extended five view-independent landmarks (e.g.\ the aortic arch summit)
to the missing projection, and revised the rules of the 52 definitions whose placement
the reviewers corrected. Eight angles and axes (e.g.\ costophrenic, tracheal axis) have a
vertex and a point per limb. The
final list has \nfinal{} definitions and $|\mathcal{V}|=\nslots{}$ class--landmark
pairs with $K_{\max}=\nkmax{}$ (Fig.~\ref{fig:atlas}).

\noindent\textbf{Silver annotations.} Since mask predictions on real CXRs might not offer reliable masks, we derive the masks from the PAXRay++ dataset which projects
its 177 structures from CT. The rules on its masks yield silver landmarks for training.

\begin{table}[b]
  \centering
  \caption{Landmark detection and segmentation on PAXRay++ test (silver) and 50 MIMIC-CXR
  images (gold; inter-reader OKS tolerances) [95\% bootstrap CI]. MRE in px, lower is better. }
  \label{tab:main}
  \setlength{\tabcolsep}{1.5pt}
  \footnotesize
  \resizebox{\linewidth}{!}{%
  \begin{tabular}{lcccccc}
    \toprule
    & \multicolumn{3}{c}{Silver} & \multicolumn{3}{c}{Gold (real)} \\
    \cmidrule(lr){2-4}\cmidrule(lr){5-7}
    Method & OKS-AP & PCK & Seg.\,AP & OKS-AP & PCK & MRE \\
    \midrule
    \multicolumn{7}{l}{\emph{Rules on predicted masks}} \\
    \hspace{2pt}CXAS~\cite{cxas}     & 32.3\,{\tiny[31.6,33.2]} & 76.1\,{\tiny[75.0,77.1]} & 48.0\,{\tiny[47.0,49.1]} & 38.9\,{\tiny[35.0,45.7]} & 63.5\,{\tiny[58.6,68.8]} & 40.7\,{\tiny[27.1,58.9]} \\
    \hspace{2pt}Seg.\ only (dense)   & 30.4\,{\tiny[30.0,31.2]} & 78.2\,{\tiny[77.3,79.0]} & \textbf{53.1}\,{\tiny[52.3,53.9]} & 44.3\,{\tiny[41.3,50.3]} & 74.7\,{\tiny[70.7,78.7]} & 15.6\,{\tiny[13.2,18.7]} \\
    \multicolumn{7}{l}{\emph{Landmark only}} \\
    \hspace{2pt}HRNet-W32~\cite{hrnet}   & 32.7\,{\tiny[32.2,33.6]} & 76.1\,{\tiny[75.2,76.9]} & -- & 46.4\,{\tiny[42.3,53.0]} & 71.2\,{\tiny[65.9,76.0]} & 17.1\,{\tiny[14.5,20.2]} \\
    \hspace{2pt}ViTPose-B~\cite{vitpose} & 33.4\,{\tiny[32.9,34.3]} & 77.1\,{\tiny[76.3,77.9]} & -- & 45.3\,{\tiny[41.4,51.5]} & 71.2\,{\tiny[66.4,76.0]} & 16.7\,{\tiny[14.7,19.6]} \\
    \hspace{2pt}RAD-DINO (dense)         & 36.1\,{\tiny[35.6,37.0]} & 79.8\,{\tiny[79.0,80.6]} & -- & 50.8\,{\tiny[47.3,56.9]} & 78.5\,{\tiny[74.9,81.9]} & 14.2\,{\tiny[12.4,16.5]} \\
    \midrule
    \multicolumn{7}{l}{\emph{Joint mask + landmark}} \\
    \hspace{2pt}Mask R-CNN~\cite{maskrcnn} & 25.9\,{\tiny[25.5,26.7]} & 71.6\,{\tiny[70.8,72.4]} & 37.0\,{\tiny[36.5,37.6]} & 45.1\,{\tiny[42.5,50.3]} & 73.7\,{\tiny[69.8,77.5]} & 15.5\,{\tiny[13.6,18.0]} \\
    \hspace{2pt}Query-based      & 35.0\,{\tiny[34.5,35.8]} & 79.3\,{\tiny[78.3,80.0]} & 52.0\,{\tiny[51.3,52.8]} & 50.1\,{\tiny[46.7,56.2]} & \textbf{79.2}\,{\tiny[75.7,82.6]} & \textbf{14.0}\,{\tiny[12.3,16.1]} \\
    \hspace{2pt}Dense            & \textbf{36.8}\,{\tiny[36.2,37.7]} & \textbf{80.6}\,{\tiny[79.7,81.4]} & 52.4\,{\tiny[51.7,53.3]} & \textbf{51.1}\,{\tiny[47.7,56.6]} & 76.7\,{\tiny[72.8,80.0]} & 14.2\,{\tiny[12.5,16.3]} \\
    \midrule
    Reader vs.\ reader & -- & -- & -- & 50.6 & 78.3 & 13.6 \\
    \bottomrule
  \end{tabular}}
\end{table}

\subsection{Gold standard and landmark-specific tolerances}
\label{ssec:gold}

Two readers with a medical background annotated 50 MIMIC-CXR images (30 frontal and 20
lateral from  42 studies of 42 patients, none in any training set), of which 26 were drawn at
random, 16 from report-positive studies of the diagnostic cohorts and 8 where learned and
rule-derived landmarks disagree most. They read nine landmark groups (trachea, carina,
oesophagogastric junction, cavoatrial junction, diaphragm, aortic arch, tracheobronchial
angle, scapulae, sternum) group by group, starting from points
derived by the rules from the masks of a segmentation model. They moved 76\% of these points and marked 15\% as not visible. The gold
position is the mean of both readers. Landmarks either reader marks as not visible are
excluded (visibility agreement 84\%), which leaves the oesophagogastric junction unscored.
The readers differ by a median of 7.5 px at $512{\times}512$.
Since a vertebral corner can be placed far more precisely than a hilum, we evaluate
with the object keypoint similarity~\cite{coco} and a tolerance per landmark:
\begin{equation}
  \mathrm{OKS}_j=\frac{\sum_k \exp\!\big(-\|\hat{\mathbf{p}}_{jk}-\mathbf{p}_{jk}\|^2/(2s_j^2\kappa_{y_jk}^2)\big)\,v_{jk}}{\sum_k v_{jk}},
  \label{eq:oks}
\end{equation}
where $s_j^2$ is the area of $M_j$ (on the gold standard, of the predicted mask the readers
saw) and, following COCO, $\kappa_{yk}=2\sigma_{yk}$ with $\sigma_{yk}$ the scale-normalised
inter-reader standard deviation of landmark $(y,k)$. With $d$ the distance between the readers,
each deviates from their mean by $d/2$, so $\sigma_{yk}^2=\mathbb{E}[d^2]/(4s^2)$ (median
0.10, from 0.04 for the left hemidiaphragm dome to 0.23 for the gastric fundus apex). On silver
labels, $\sigma_{yk}{=}0.05$ for all landmarks. OKS-AP is the COCO average precision over OKS
thresholds $0.50{:}0.05{:}0.95$, averaged over structure classes. We also report PCK (within 0.1 of
the box diagonal. Missed structures count as wrong) and mean radial error (MRE, px at $512{\times}512$).

\begin{table*}[b]
  \centering
  \caption{Diagnostic potential on MIMIC-CXR: AUROC [95\% bootstrap CI] against report-derived
  labels for rules on CXAS masks and landmarks of the joint dense model (Dense). Failed
  measurements count as wrong. $\Delta$: Dense$\,-\,$Rules (CXAS), paired CI; $^\ast$: significant (paired
  DeLong, Holm over 23 comparisons). Marked Bold means better where significant. $^\dagger$: ${<}$50
  positives. F+L: frontal and lateral. }
  \label{tab:usecases}
  \setlength{\tabcolsep}{2.2pt}
  \footnotesize
  \begin{tabular}{l|ccc|ccc|ccc}
    \toprule
    & \multicolumn{3}{c|}{PA} & \multicolumn{3}{c|}{AP} & \multicolumn{3}{c}{Lateral} \\
    Finding: criterion & \shortstack{Rules\\(CXAS)} & Dense & $\Delta$ & \shortstack{Rules\\(CXAS)} & Dense & $\Delta$ & \shortstack{Rules\\(CXAS)} & Dense & $\Delta$ \\
    \midrule
    \multicolumn{10}{l}{\emph{Findings}} \\
    Cardiomegaly: CTR~\cite{danzer1919} & .92\,{\tiny[.89,.94]} & .93\,{\tiny[.90,.95]} & +.01\,{\tiny[$-$.01,+.02]} & .80\,{\tiny[.76,.84]} & \textbf{.90}\,{\tiny[.88,.93]} & +.10$^\ast$\,{\tiny[+.07,+.13]} & -- & -- & -- \\
    Cardiomegaly: Hoffman--Rigler sign~\cite{hoffman1965} & -- & -- & -- & -- & -- & -- & .11\,{\tiny[.08,.13]} & \textbf{.66}\,{\tiny[.61,.70]} & +.55$^\ast$\,{\tiny[+.50,+.60]} \\
    Hyperinflation: rib level of dome~\cite{eur16260_1996} & .75\,{\tiny[.72,.79]} & .79\,{\tiny[.75,.83]} & +.03\,{\tiny[+.01,+.06]} & .63\,{\tiny[.58,.68]} & \textbf{.76}\,{\tiny[.72,.80]} & +.13$^\ast$\,{\tiny[+.10,+.17]} & -- & -- & -- \\
    Hyperinflation: dome flattening~\cite{pratt1987} & -- & -- & -- & -- & -- & -- & .59\,{\tiny[.55,.63]} & .61\,{\tiny[.57,.65]} & +.02\,{\tiny[$-$.04,+.07]} \\
    Pleural effusion: CP-angle blunting~\cite{blackmore1996} & .59\,{\tiny[.55,.64]} & \textbf{.82}\,{\tiny[.79,.86]} & +.23$^\ast$\,{\tiny[+.18,+.28]} & .47\,{\tiny[.42,.52]} & \textbf{.70}\,{\tiny[.66,.75]} & +.23$^\ast$\,{\tiny[+.17,+.29]} & -- & -- & -- \\
    Pleural effusion: posterior sulcus~\cite{collins1972} & -- & -- & -- & -- & -- & -- & .54\,{\tiny[.49,.58]} & \textbf{.73}\,{\tiny[.69,.77]} & +.19$^\ast$\,{\tiny[+.14,+.25]} \\
    Elevated hemidiaphragm: dome asymmetry & .83\,{\tiny[.80,.87]} & .87\,{\tiny[.84,.90]} & +.04\,{\tiny[+.01,+.07]} & .68\,{\tiny[.64,.73]} & \textbf{.79}\,{\tiny[.75,.82]} & +.10$^\ast$\,{\tiny[+.06,+.14]} & -- & -- & -- \\
    Scoliosis: spine-axis Cobb angle~\cite{cobb1948} & .73\,{\tiny[.69,.77]} & \textbf{.84}\,{\tiny[.81,.87]} & +.11$^\ast$\,{\tiny[+.07,+.15]} & .63\,{\tiny[.58,.67]} & \textbf{.84}\,{\tiny[.81,.87]} & +.21$^\ast$\,{\tiny[+.16,+.26]} & -- & -- & -- \\
    Kyphosis: spine-axis angle T4--T12~\cite{fon1980} & -- & -- & -- & -- & -- & -- & .01\,{\tiny[.01,.01]} & \textbf{.88}\,{\tiny[.86,.90]} & +.87$^\ast$\,{\tiny[+.85,+.89]} \\
    Pectus excavatum: Haller index (F+L)~\cite{haller1987,khanna2010} & .43\,{\tiny[.38,.48]} & \textbf{.78}\,{\tiny[.74,.82]} & +.35$^\ast$\,{\tiny[+.29,+.41]} & .51\,{\tiny[.36,.64]} & .63\,{\tiny[.51,.75]} & +.12$^\dagger$\,{\tiny[$-$.08,+.34]} & -- & -- & -- \\
    Vertebral fracture: height loss~\cite{genant1993} & -- & -- & -- & -- & -- & -- & .01\,{\tiny[.01,.02]} & \textbf{.63}\,{\tiny[.60,.66]} & +.61$^\ast$\,{\tiny[+.58,+.64]} \\
    Tortuous aorta: aortic knob width~\cite{tomita2020} & .74\,{\tiny[.70,.78]} & .74\,{\tiny[.70,.78]} & .00\,{\tiny[$-$.02,+.02]} & .57\,{\tiny[.52,.62]} & .59\,{\tiny[.55,.64]} & +.02\,{\tiny[$-$.01,+.05]} & -- & -- & -- \\
    \midrule
    \rowcolor{gray!12}\multicolumn{10}{l}{\emph{Acquisition quality}} \\
    \rowcolor{gray!12}Inspiration: rib level of dome~\cite{eur16260_1996} & .90\,{\tiny[.88,.92]} & .88\,{\tiny[.86,.91]} & $-$.02\,{\tiny[$-$.04,.00]} & .86\,{\tiny[.83,.89]} & .86\,{\tiny[.82,.88]} & $-$.01\,{\tiny[$-$.03,+.02]} & -- & -- & -- \\
    \rowcolor{gray!12}Rotation: clavicle--midline symmetry~\cite{eur16260_1996,marais2023} & .78\,{\tiny[.73,.81]} & .76\,{\tiny[.71,.80]} & $-$.02\,{\tiny[$-$.05,+.01]} & .73\,{\tiny[.69,.77]} & .74\,{\tiny[.70,.79]} & +.01\,{\tiny[$-$.03,+.05]} & -- & -- & -- \\
    \midrule
    Mean & .74\,{\tiny[.73,.75]} & .82\,{\tiny[.81,.83]} & +.08\,{\tiny[+.07,+.09]} & .65\,{\tiny[.63,.67]} & .76\,{\tiny[.74,.77]} & +.10\,{\tiny[+.08,+.13]} & .25\,{\tiny[.24,.26]} & .70\,{\tiny[.69,.72]} & +.45\,{\tiny[+.43,+.47]} \\
    \bottomrule
  \end{tabular}
\end{table*}

\subsection{Joint mask and landmark prediction}
\label{ssec:model}

All models share a RAD-DINO~\cite{raddino} ViT-B/14 encoder, fine-tuned end-to-end, and
differ in how the two tasks are integrated.

\noindent\textbf{Single-task.} A landmark-only detector decodes the patch tokens with
deconvolutions to one heatmap per class--landmark pair ($|\mathcal{V}|=\nslots{}$
channels with stride 4) trained with an MSE loss on Gaussian targets. A segmentation-only
model uses the same decoder with one mask channel per class (stride 2), trained with BCE
and Dice.

\noindent\textbf{Joint dense.} Both heads on the same encoder apply losses at
equal weight. Landmark $(y,k)$ and  mask of class $y$ is read from its respective channel since PAXRay++ $\leq 1$ instance/class.

\noindent\textbf{Joint instance-first.} The detector first predicts instances and then
localises the landmarks of each from its features, so a landmark inherits its identity
from the predicted class. We use Mask R-CNN~\cite{maskrcnn} with a class-specific keypoint
head (each RoI reads the $K_y$ slots of its class) and a query-based segmentor~\cite{mask2former}
whose decoder embeddings predict $K_{\max}{=}\nkmax{}$ class-conditioned slots as 1-D
distributions~\cite{simcc} relative to the instance mask although no significant difference in performance to heatmap-based predictions was seen.

\section{Experiments}
\label{sec:experiments}

\noindent\textbf{Setup.} All models are trained on PAXRay++~\cite{cxas}, whose masks provide
silver labels via the rules of Sec.~\ref{ssec:taxonomy} (inconsistent side names of the
clavicles, hemidiaphragms and visible lungs reassigned by position on frontal views). We
evaluate on its test split and on the gold standard. We compare against the rules applied to masks predicted by CXAS~\cite{cxas} and by the segmentation-only
model, against the landmark-only detectors HRNet-W32~\cite{hrnet}, ViTPose-B~\cite{vitpose} and
RAD-DINO with one heatmap per class--landmark pair, and against the joint models of
Sec.~\ref{ssec:model}. Masks are scored by COCO mask AP.

\noindent\textbf{Diagnostic study.} We test whether the landmarks carry diagnostic information on
real CXRs from MIMIC-CXR~\cite{johnson2019mimic}, which none of the models saw. For each
view we sample positive and negative studies with report-derived labels~\cite{irvin2019chexpert},
implement each cited criterion as a measurement on the landmarks of the joint dense model
and compare with the same criteria computed by rules from CXAS masks. A failed measurement
counts as a wrong call.

\noindent\textbf{Implementation details.} The RAD-DINO models are fine-tuned end-to-end with
layer-wise learning-rate decay (0.75) on $512{\times}512$ inputs with RandAugment, CutMix and
elastic augmentation. Checkpoints are selected on validation OKS-AP. Dense models use
deconvolution decoders (width 512), Gaussian targets ($\sigma{=}2$) and DARK
decoding~\cite{dark}, and train for 18 epochs with AdamW (batch 8, learning rate
$5{\cdot}10^{-4}$, cosine schedule, EMA). Instance-first models use learning rate $2{\cdot}10^{-4}$.

\noindent\textbf{Landmark detection.} Rules on the segmentation-only masks reach 30.4 OKS-AP,
5.7 below the landmark-only detector with the same encoder, as mask errors propagate to the
landmarks. On the gold standard, direct prediction stays ahead of rules on the
segmentation-only masks by 6.8 {\scriptsize[2.5, 10.2]} OKS-AP and of CXAS by 12.2
{\scriptsize[6.9, 17.3]}, with an error close to that between the two readers. With 50 images, it is not separable from the landmark-only (+0.3 {\scriptsize[$-$2.1, 2.3]}) and
query-based (+1.0 {\scriptsize[$-$2.1, 3.8]}) models.

\noindent\textbf{Joint learning.} Mask supervision improves the dense detector by +0.7 {\scriptsize[+0.5, +0.9]}
OKS-AP and +0.8 {\scriptsize[+0.6, +1.0]} PCK at $-$0.7 {\scriptsize[$-$0.8, $-$0.5]} segmentation AP. The instance-first models fall below the landmark-only
detector, although the query-based model segments almost as well. We attribute this to
instance-first models reading landmarks from a query embedding or pooled region rather than a
full-image heatmap.

\noindent\textbf{Diagnostic use cases.} Table~\ref{tab:usecases} lists the criteria for which the dense model is
significantly above chance in its reference view (PA, or lateral for lateral criteria) with an
AUROC of at least 0.6. On PA images, the CTR, the CP angle, the dome asymmetry and the Cobb angle
reach 0.82--0.93. Frontal criteria are lower on AP images. The dense model exceeds the rules on CXAS masks by 0.08 {\scriptsize[0.07, 0.09]} mean
AUROC on PA, 0.10 {\scriptsize[0.08, 0.13]} on AP and 0.45 {\scriptsize[0.43, 0.47]} on lateral images. After Holm correction, the dense model is
significantly better in 12 of 23 comparisons and never significantly worse. The largest frontal gains are the
Haller index (+0.35, PA), CP-angle blunting (+0.23) and the Cobb angle (+0.11 PA, +0.21 AP). On
lateral images, the rules reach only 0.01--0.11 AUROC for the Hoffman--Rigler sign, kyphosis and vertebral
height loss, as failed measurements count as wrong and the CXAS masks lack the structures they need. 



\section{Conclusion}
\label{sec:conclusion}

We introduced a clinically curated set of \nfinal{} semantically named CXR landmarks,
defined by geometric rules and learned from silver labels on PAXRay++. Learned jointly with
masks in a shared dense model, they improve at a small segmentation cost, while instance-first
models lose this benefit. On real CXRs, direct prediction stays ahead of rules on predicted
masks, with an error close to the difference between two readers, and the landmarks carry
diagnostic information across 14 criteria, most clearly on lateral views.

\section{Compliance with ethical standards}
\label{sec:ethics}

This retrospective study used open-access PAXRay++~\cite{cxas} and credentialed-access,
de-identified MIMIC-CXR~\cite{johnson2019mimic} data. Ethical approval was not required under
their licences.

\section{Acknowledgments}
\label{sec:acknowledgments}

Constantin Seibold was supported by the Medical Data Scientist Program of the
Medical Faculty Heidelberg. Claude Code (Anthropic) was used for literature search, for implementing the code, and for drafting and editing the text of all sections. The authors reviewed all output and take full responsibility for the content. The authors have no conflicts of interest to declare. 

\bibliographystyle{IEEEbib}
\bibliography{strings,refs,anatomy}

@article{danzer1919, title={The cardiothoracic ratio: an index of cardiac enlargement}, volume={157}, DOI={10.1097/00000441-191904000-00007}, number={4}, journal={Am. J. Med. Sci.}, author={Danzer, C. Saul}, year={1919}, pages={513--554} }

@article{collins1972, title={Minimal Detectable Pleural Effusions: A Roentgen Pathology Model}, volume={105}, DOI={10.1148/105.1.51}, number={1}, journal={Radiology}, author={Collins, James D. and Burwell, Douglas and Furmanski, Stanley and Lorber, Peter and Steckel, Richard J.}, year={1972}, pages={51--53} }

@article{blackmore1996, title={Pleural fluid volume estimation: A chest radiograph prediction rule}, volume={3}, DOI={10.1016/s1076-6332(05)80373-3}, number={2}, journal={Acad. Radiol.}, author={Blackmore, C. Craig and Black, William C. and Dallas, Robert V. and Crow, Harte C.}, year={1996}, pages={103--109} }

@book{eur16260_1996,
  author    = {{European Commission}},
  title     = {European Guidelines on Quality Criteria for Diagnostic Radiographic Images},
  publisher = {Office for Official Publications of the European Communities},
  address   = {Luxembourg},
  year      = {1996}
}

@article{hansell2008, title={Fleischner Society: Glossary of Terms for Thoracic Imaging}, volume={246}, DOI={10.1148/radiol.2462070712}, number={3}, journal={Radiology}, author={Hansell, David M. and Bankier, Alexander A. and MacMahon, Heber and McLoud, Theresa C. and M{\"u}ller, Nestor L. and Remy, Jacques}, year={2008}, pages={697--722} }

@article{pratt1987, title={Role of conventional chest radiography in diagnosis and exclusion of emphysema}, volume={82}, DOI={10.1016/0002-9343(87)90163-x}, number={5}, journal={Am. J. Med.}, author={Pratt, Philip C.}, year={1987}, pages={998--1006} }

@article{tomita2020, title={Aortic Knob Width as a Novel Indicator of Atherosclerosis and Obstructive Sleep Apnea}, volume={27}, DOI={10.5551/jat.50286}, number={6}, journal={J. Atheroscler. Thromb.}, author={Tomita, Yasuhiro and Kasai, Takatoshi and Ishiwata, Sayaki and Daida, Hiroyuki and Narui, Koji}, year={2020}, pages={501--508} }

@article{goodman1976, title={Radiographic evaluation of endotracheal tube position}, volume={127}, DOI={10.2214/ajr.127.3.433}, number={3}, journal={Am. J. Roentgenol.}, author={Goodman, L. R. and Conrardy, P. A. and Laing, F and Singer, MM}, year={1976}, pages={433--434} }

@article{baskin2008, title={Cavoatrial Junction and Central Venous Anatomy: Implications for Central Venous Access Tip Position}, volume={19}, DOI={10.1016/j.jvir.2007.09.005}, number={3}, journal={J. Vasc. Interv. Radiol.}, author={Baskin, Kevin M. and Jimenez, Rafael M. and Cahill, Anne Marie and Jawad, Abass F. and Towbin, Richard B.}, year={2008}, pages={359--365} }

@article{marais2023, title={Patient rotation chest X-rays and the consequences of misinterpretation in paediatric radiology}, volume={47}, DOI={10.1016/j.prrv.2023.05.003}, journal={Paediatr. Respir. Rev.}, author={Marais, Jani and Venkatakrishna, Shyam Sunder B. and Calle-Toro, Juan S. and Goussard, Pierre and Eber, Ernst and Andronikou, Savvas}, year={2023}, pages={41--50} }

@incollection{cobb1948,
  author    = {Cobb, J. R.},
  title     = {Outline for the study of scoliosis},
  booktitle = {AAOS Instructional Course Lectures},
  volume    = {5},
  pages     = {261--275},
  publisher = {J. W. Edwards},
  address   = {Ann Arbor},
  year      = {1948}
}

@article{genant1993, title={Vertebral fracture assessment using a semiquantitative technique}, volume={8}, DOI={10.1002/jbmr.5650080915}, number={9}, journal={J. Bone Miner. Res.}, author={Genant, Harry K. and Wu, Chun Y. and van Kuijk, Cornelis and Nevitt, Michael C.}, year={1993}, pages={1137--1148} }

@inproceedings{hrnet, title={Deep High-Resolution Representation Learning for Human Pose Estimation}, DOI={10.1109/cvpr.2019.00584}, booktitle={Proc. CVPR}, author={Sun, Ke and Xiao, Bin and Liu, Dong and Wang, Jingdong}, year={2019}, pages={5686--5696} }

@inproceedings{mask2former, title={Masked-attention Mask Transformer for Universal Image Segmentation}, DOI={10.1109/cvpr52688.2022.00135}, booktitle={Proc. CVPR}, author={Cheng, Bowen and Misra, Ishan and Schwing, Alexander G. and Kirillov, Alexander and Girdhar, Rohit}, year={2022}, pages={1280--1289} }

@inproceedings{coco, title={Microsoft {COCO}: Common Objects in Context}, DOI={10.1007/978-3-319-10602-1_48}, booktitle={Proc. ECCV}, author={Lin, Tsung-Yi and Maire, Michael and Belongie, Serge and Hays, James and Perona, Pietro and Ramanan, Deva and Doll{\'a}r, Piotr and Zitnick, C. Lawrence}, year={2014}, pages={740--755} }

@article{raddino, title={Exploring scalable medical image encoders beyond text supervision}, volume={7}, DOI={10.1038/s42256-024-00965-w}, number={1}, journal={Nat. Mach. Intell.}, author={P{\'e}rez-Garc{\'i}a, Fernando and Sharma, Harshita and Bond-Taylor, Sam and Bouzid, Kenza and Salvatelli, Valentina and Ilse, Maximilian and Bannur, Shruthi and Castro, Daniel C. and Schwaighofer, Anton and Lungren, Matthew P. and Wetscherek, Maria Teodora and Codella, Noel and Hyland, Stephanie L. and Alvarez-Valle, Javier and Oktay, Ozan}, year={2025}, pages={119--130} }

@article{chexmask, title={{CheXmask}: a large-scale dataset of anatomical segmentation masks for multi-center chest x-ray images}, volume={11}, DOI={10.1038/s41597-024-03358-1}, number={1}, journal={Sci. Data}, pages={511}, author={Gaggion, Nicol{\'a}s and Mosquera, Candelaria and Mansilla, Lucas and Saidman, Julia Mariel and Aineseder, Martina and Milone, Diego H. and Ferrante, Enzo}, year={2024} }

@inproceedings{simcc, title={{SimCC}: A Simple Coordinate Classification Perspective for~Human Pose Estimation}, DOI={10.1007/978-3-031-20068-7_6}, booktitle={Proc. ECCV}, author={Li, Yanjie and Yang, Sen and Liu, Peidong and Zhang, Shoukui and Wang, Yunxiao and Wang, Zhicheng and Yang, Wankou and Xia, Shu-Tao}, year={2022}, pages={89--106} }

@inproceedings{dark, title={Distribution-Aware Coordinate Representation for Human Pose Estimation}, DOI={10.1109/cvpr42600.2020.00712}, booktitle={Proc. CVPR}, author={Zhang, Feng and Zhu, Xiatian and Dai, Hanbin and Ye, Mao and Zhu, Ce}, year={2020}, pages={7091--7100} }

@misc{cxas,
howpublished = {arXiv:2306.03934},
doi = {10.48550/ARXIV.2306.03934},
author = {Seibold, Constantin and Jaus, Alexander and Fink, Matthias A. and Kim, Moon and Rei{\ss}, Simon and Herrmann, Ken and Kleesiek, Jens and Stiefelhagen, Rainer},
title = {Accurate Fine-Grained Segmentation of Human Anatomy in Radiographs via Volumetric Pseudo-Labeling},
year = {2023}
}

@inproceedings{vitpose,
booktitle = {Proc. NeurIPS},
doi = {10.48550/ARXIV.2204.12484},
author = {Xu, Yufei and Zhang, Jing and Zhang, Qiming and Tao, Dacheng},
title = {{ViTPose}: Simple Vision Transformer Baselines for Human Pose Estimation},
year = {2022}
}

@inproceedings{paxray,
  author    = {Seibold, Constantin and Rei{\ss}, Simon and Sarfraz, M. Saquib and Fink, Matthias A. and Mayer, Victoria and Sellner, Jan and Kim, Moon Sung and Maier-Hein, Klaus H. and Kleesiek, Jens and Stiefelhagen, Rainer},
  title     = {Detailed Annotations of Chest {X}-Rays via {CT} Projection for Report Understanding},
  booktitle = {Proc. BMVC},
  year      = {2022}
}

@article{johnson2019mimic, title={{MIMIC-CXR}, a de-identified publicly available database of chest radiographs with free-text reports}, volume={6}, DOI={10.1038/s41597-019-0322-0}, number={1}, journal={Sci. Data}, author={Johnson, Alistair E. W. and Pollard, Tom J. and Berkowitz, Seth J. and Greenbaum, Nathaniel R. and Lungren, Matthew P. and Deng, Chih-ying and Mark, Roger G. and Horng, Steven}, year={2019}, pages={317} }

@inproceedings{irvin2019chexpert, title={{CheXpert}: A large chest radiograph dataset with uncertainty labels and expert comparison}, booktitle={Proc. AAAI}, volume={33}, pages={590--597}, DOI={10.1609/aaai.v33i01.3301590}, author={Irvin, Jeremy and Rajpurkar, Pranav and Ko, Michael and Yu, Yifan and Ciurea-Ilcus, Silviana and Chute, Chris and Marklund, Henrik and Haghgoo, Behzad and Ball, Robyn and Shpanskaya, Katie and others}, year={2019} }

@article{hoffman1965, title={Evaluation of left ventricular enlargement in the lateral projection of the chest}, volume={85}, DOI={10.1148/85.1.93}, number={1}, journal={Radiology}, author={Hoffman, Richard B. and Rigler, Leo G.}, year={1965}, pages={93--100} }

@article{fon1980, title={Thoracic kyphosis: range in normal subjects}, volume={134}, DOI={10.2214/ajr.134.5.979}, number={5}, journal={Am. J. Roentgenol.}, author={Fon, Gregory T. and Pitt, Michael J. and Thies, A. Craig}, year={1980}, pages={979--983} }

@article{haller1987, title={Use of {CT} scans in selection of patients for pectus excavatum surgery: a preliminary report}, volume={22}, DOI={10.1016/S0022-3468(87)80585-7}, number={10}, journal={J. Pediatr. Surg.}, author={Haller, J. Alex and Kramer, S. S. and Lietman, S. A.}, year={1987}, pages={904--906} }

@article{khanna2010, title={Comparison of {Haller} index values calculated with chest radiographs versus {CT} for pectus excavatum evaluation}, volume={40}, number={11}, journal={Pediatr. Radiol.}, author={Khanna, Geetika and Jaju, Alok and Don, Steven and Keys, Tiffany and Hildebolt, Charles F.}, year={2010}, pages={1763--1767} }

@inproceedings{maskrcnn, title={{Mask R-CNN}}, DOI={10.1109/iccv.2017.322}, booktitle={Proc. ICCV}, author={He, Kaiming and Gkioxari, Georgia and Doll{\'a}r, Piotr and Girshick, Ross}, year={2017}, pages={2980--2988} }

\end{document}